\documentclass[10pt,twocolumn,letterpaper]{article}

\usepackage[pagenumbers]{cvpr} 

\usepackage{graphicx}
\usepackage{amsmath}
\usepackage{amssymb}
\usepackage{booktabs}

\usepackage[pagebackref,breaklinks,colorlinks]{hyperref}

\usepackage[capitalize]{cleveref}
\crefname{section}{Sec.}{Secs.}
\Crefname{section}{Section}{Sections}
\Crefname{table}{Table}{Tables}
\crefname{table}{Tab.}{Tabs.}

\def\cvprPaperID{29} 
\def\confName{CVPR}
\def\confYear{2022}

\begin{document}

\title{Facial Expression Classification using Fusion of Deep Neural Network in Video for the 3rd ABAW3 Competition}

\author{Kim Ngan Phan\\
Dept. of Artificial Intelligence Convergence\\
Chonnam National University\\
Gwangju, South Korea \\
{\tt\small kimngan260997@gmail.com}
\and
Hong-Hai Nguyen\\
Dept. of Artificial Intelligence Convergence\\
Chonnam National University\\
Gwangju, South Korea \\
{\tt\small honghaik14@gmail.com}
\and
Van-Thong Huynh\\
Dept. of Artificial Intelligence Convergence\\
Chonnam National University\\
Gwangju, South Korea \\
{\tt\small vthuynh@jnu.ac.kr}
\and
Soo-Hyung Kim\thanks{Corresponding author}\\
Dept. of Artificial Intelligence Convergence\\
Chonnam National University\\
Gwangju, South Korea \\
{\tt\small shkim@jnu.ac.kr}
}
\maketitle

\begin{abstract}
For computers to recognize human emotions, expression classification is an equally important problem in the human-computer interaction area. In the 3rd Affective Behavior Analysis In-The-Wild competition, the task of expression classification includes eight classes with six basic expressions of human faces from videos. In this paper, we employ a transformer mechanism to encode the robust representation from the backbone. Fusion of the robust representations plays an important role in the expression classification task. Our approach achieves 30.35\%  and 28.60\% for the $F_1$ score on the validation set and the test set, respectively. This result shows the effectiveness of the proposed architecture based on the Aff-Wild2 dataset.
\end{abstract}

\section{INTRODUCTION}
\label{sec:intro}
Understanding Affective Behavior is playing an essential role in the interaction between computers and humans \cite{kollias2021analysing}. This interaction makes it possible for computers to understand human behaviors and emotions and feelings. For many years, scientists have been working to build an intelligent and automated machine that can understand and serve humans in many fields of health, education, and services. The emotion recognition system allows for receiving many different data sources such as biological signals, visuals, or documents. Visual data directly depicts interpretations of emotions through facial expressions, thus playing an important role in emotion classification. In 1969, Ekman proposed six basic emotions in \cite{ekman1969pan} including anger, disgust, fear, happiness, sadness, and surprise. They are used popular but are not sufficient to express complex human emotional states. In 2022, the 3rd Affective Behavior Analysis In-The-Wild (ABAW) Competition is organized with the major goal of building system and improving the emotional recognition ability of machines \cite{kollias2022abaw, kollias2021analysing, kollias2020analysing, kollias2021distribution, kollias2021affect, kollias2019expression, kollias2019face, kollias2019deep, zafeiriou2017aff}. The competition consists of 4 challenges: valence-arousal estimation, expression classification, action unit (AU) detection, multi-task-learning. Each task corresponds to huge datasets with different sizes from Aff-Wild2 Database \cite{kollias2022abaw, kollias2021analysing, kollias2020analysing, kollias2021distribution, kollias2021affect, kollias2019expression, kollias2019face, kollias2019deep, zafeiriou2017aff}. They include videos and cropped and aligned frames that contain annotations in terms. We perform the expression classification task in this paper. In \cite{gera2020affect}, the author proposes the pre-trained ResNet model that combines spatio-channel attention and efficient channel attention to get robust features. In \cite{thinh2021emotion}, the authors propose the multi-task learning technique for the pre-trained Resnet model, and they use Focal Loss to solve the imbalanced dataset. In this study, we propose the combination of representative features from deep learning networks for the expression classification task. We opt RegNet as the backbone of our network. The transformer encoder plays a role as the embedding layer to extract robust representations from the backbone. We employ multi-head attention with the space of the attention heads being expanded whose embed dimensions are flexible. Our model archives better performance than the baseline on the validation set. Section 2 describes the proposed method. The training details and results are reported at the last of the paper.

\section{PROPOSED METHOD}
\label{sec:formatting}
\begin{figure}
  \begin{center}
  \includegraphics[width=6cm]{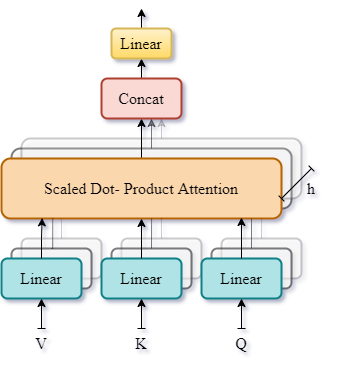}
  \caption{Overview of Multi-Head attention Module.}
  \label{fig:att}
  \end{center}
\end{figure}

\begin{figure}
  \begin{center}
  \includegraphics[width=9cm]{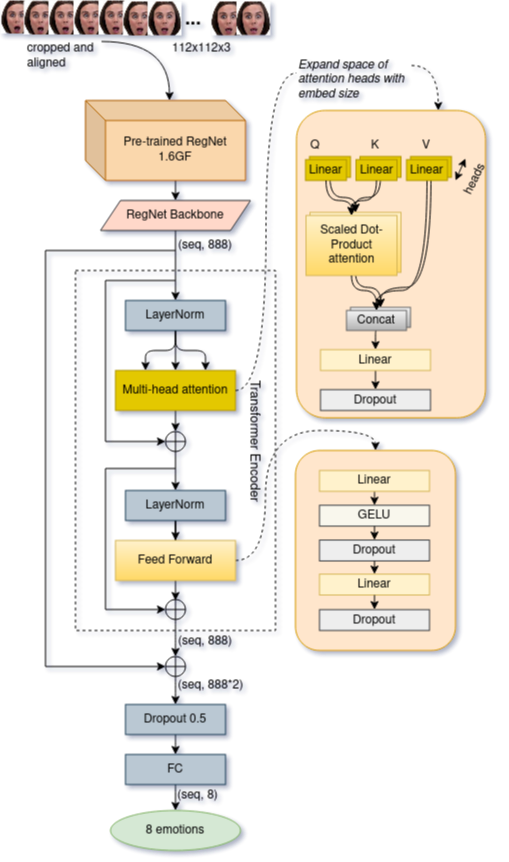}
  \caption{Detail of our architecture for facial expression classification.}
  \label{fig:model}
  \end{center}
\end{figure}

In \cite{vaswani2017attention}, the authors introduce multi-head attention has several attention layers in parallel. The output of attention layers is a weighted sum of the value. In multi-head attention, the attention layer runs independently and their outputs are concatenated and linearly transformed into the new space with the expected dimension.  Fig \ref{fig:att} is overview of multi-head attention. The authors recommend the Transformer architecture that has the encoder-decoder structure to build global dependencies for the connection of input and output. The transformer encoder including a stack of multi-head attention mechanisms and feed-forward networks can map an input sequence to representation features. In this paper, we employ a transformer encoder with multi-head attention as the embedded layer to generate sequence representations. In the multi-head attention, the heads are expanded with flexible embed dimensions to enhance the information on the heads. The transformer helps encode the robust representations for the backbone of the model. We also employ the pre-trained model RegNet \cite{radosavovic2020designing} as the backbone for the proposed network. We use the weight of RegNet with 1.6GF architecture on ImageNet dataset \cite{deng2009imagenet} to extract feature of images. The pre-trained model takes the input size of the image as 112x112x3. The backbone is extracted with 888 features by the flattened layer of the pre-trained model. We reshape the backbone to (batch size, sequence, feature) and fed it into the transformer encoder. In this work, we opt for the length of the sequence to be 64. To improve the performance of the classification task, we consider the fusion of the robust representations before entering the classification model. We combine the backbones and the representation of the transformer encoder mechanism to get the final representation features for the expression classification task. Dropout layer with 50\% information and dense layers of 8 neurons are used for final output corresponding to 8 expressions of human. Fig  \ref{fig:model} describe detail of our architecture. 

\section{EXPERIMENTS}
\subsection{Dataset}
The large-scale in-the-wild Aff-Wild2 database contains 548 videos with approximately 2.7 million frames. The training process is conducted on the Aff-Wild2 database. Data is used containing annotations on 6 basic expressions including Anger, Disgust, Fear, Happiness, Sadness, Surprise, plus Neutral state, and Other which denotes emotional expressions other than the 6 basic states. The participants are provided with frames in RGB color space from the database. The images are cropped and aligned with an input size of 112x112 from the video.
\subsection{Training Details}
The network is trained on the Pytorch framework. We use Adam optimization \cite{kingma2014adam} to update the weights. We opt Focal Loss \cite{lin2017focal} for the classification task of eight emotions. The training process automatically finds the best learning rate. The batch size of 16 is trained during the training process. Our model learns the epoch of 30 and saves the best performance on the validation set. In this work, the final result is evaluated across the average $F_1$ score of 8 emotion categories:

$ F_1^{final} = \frac{\sum{F_1^{expr}}}{8} $ where $F_1^{expr}$ is $F_1$ score of each expression.

\begin{table}
  \centering
  \begin{tabular}{|c|c|c|}
    \toprule
    Parameters & Transformer Encoder\\
    \midrule
    Depth & 1 \\
    Head Number &  2\\
    Embedded Dimension & 64\\
    Feedforward Dimension & 512 \\
    Dropout Value & 0.\\ 
    \bottomrule
  \end{tabular}
  \caption{Hyperparameter of our architecture.}
  \label{tab:parameter}
\end{table}

\begin{table}
  \centering
  \scalebox{1.0}
  {
  \begin{tabular}{|c|c|c|}
    \toprule
    Model  &  Validation & Test \\
    \midrule
    Baseline (VGG16) \cite{kollias2022abaw} & 23 & 20.50\\
    Our Model (Attention) & 29.11 & 26.32\\
    \textbf{Our Model} (Transformer)  &  \textbf{30.35} & \textbf{28.60} \\
    \bottomrule
  \end{tabular}
  }
  \caption{Expression classification results of our model on the validation set and the test set.}
  \label{tab:results}
\end{table}

\subsection{Results}
We report results by $F_1$ score in table \ref{tab:results} on both the validation set and the test set. In the baseline \cite{kollias2022abaw}, the authors perform the pre-trained VGG16 network on the VGGFACE dataset and get softmax probabilities for the 8 expression predictions. They archive the $F_1$ score of 23\% and 20.50\% on the validation set and the test set, respectively. In the proposed model, we not only use the backbone representation but also combine the additional representation of the transformer encoder. As a result, this fusion has a lot of information so that the dropout layer is applied immediately after. Our model performs better than the baseline. We try combining the backbone representation and representation using multi-head attention mechanism. This fusion doesn't perform better than the transformer mechanism. This shows the effectiveness of the transformer as an embedding layer to encode the salient information of the backbone.

\section{CONCLUSION}
In the 3rd Affective Behavior Analysis In-The-Wild (ABAW3) Competition, we have the opportunity to contribute research results in the field of human-computer interaction. Our proposed model performs the expression classification task based on videos of the Aff-Wild2 database. We recommend the fusion of the robust representative features from deep neural layer branches including the pre-trained RegNet model and transformer encoder. Results show the effectiveness of fusion for facial expression classification task.

\section*{Acknowledgments}
This work was supported by the National Research Foundation of Korea (NRF) grant funded by the Korea government (MSIT) (NRF-2020R1A4A1019191) and Basic Science Research Program through the National Research Foundation of Korea (NRF) funded by the Ministry of Education (NRF-2021R1I1A3A04036408).

{\small
\bibliographystyle{ieee_fullname}
\bibliography{egbib}

}

\end{document}